\documentclass[10pt, a4paper]{article}
\usepackage{lrec}

\usepackage{graphicx, color}
\usepackage{tabularx}
\usepackage{soul}
\usepackage{booktabs}
\usepackage{tipa}
\usepackage{multirow}
\usepackage{accents}

\usepackage{epstopdf}
\usepackage{amssymb}
\usepackage{pifont}
\newcommand{\cmark}{\ding{51}}%
\newcommand{\xmark}{\ding{55}}%

\usepackage{hyperref}
\usepackage{xstring}
\usepackage{xspace}

\newcommand*\underdot[1]{%
  \underaccent{\dot}{#1}}

\newcommand{\Ni}{({\em i})~}
\newcommand{\Nii}{({\em ii})~}
\newcommand{\Niii}{({\em iii})~}

\newcommand*{\yoruba}{Yor\`ub\'a\xspace}

\title{Massive vs. Curated Embeddings for Low-Resourced Languages:\\ the Case of \yoruba and Twi}

\name{%
Jesujoba O. Alabi$^{\star \dagger \ddagger}$\thanks{($\star$) Equal contribution to the work, author names are arranged alphabetically by last name.} 
Kwabena Amponsah-Kaakyire$^{\star \dagger \ddagger}$  
David I. Adelani$^{\ddagger \parallel}$
Cristina Espa\~na-Bonet$^{\dagger \ddagger}$  
}

\address{$^\dagger$DFKI GmbH, Saarbr\"ucken, Germany \\
$^\parallel$Spoken Language Systems (LSV), Saarland Informatics Campus, 
$^\ddagger$Saarland University, Saarbr\"ucken, Germany \\
         \{jesujoba\_oluwadara.alabi, kwabena.amponsah{-}kaakyire, cristinae\}@dfki.de, didelani@lsv.uni-saarland.de\\
        }

\abstract{
The success of several architectures to learn semantic representations from unannotated text and the availability of these kind of texts in online multilingual resources such as Wikipedia 
has facilitated the massive and automatic creation of resources for multiple languages. The evaluation of such resources is usually done for the high-resourced languages, where one has a smorgasbord of tasks and test sets to evaluate on. For low-resourced languages, the evaluation is more difficult and normally ignored, with the hope that the impressive capability of deep learning architectures to learn (multilingual) representations in the high-resourced setting holds in the low-resourced setting too. 
In this paper we focus on two African languages, \yoruba and Twi, and compare the word embeddings obtained in this way, with word embeddings obtained from curated corpora and a language-dependent processing. We analyse the noise in the publicly available corpora, collect high quality and noisy data for the two languages and quantify the improvements that depend not only on the amount of data but on the quality too. We also use different architectures that learn word representations both from surface forms and characters to further exploit all the available information which showed to be important for these languages. For the evaluation, we manually translate the wordsim-353 word pairs dataset from English into \yoruba and Twi. 
We extend the analysis to contextual word embeddings and evaluate multilingual BERT on a named entity recognition task. For this, we annotate with named entities the Global Voices corpus for \yoruba.
As output of the work, we provide corpora, embeddings and the test suits for both languages.
\\ 
\newline  }

\begin{document}

\maketitleabstract

\section{Introduction}
\label{s:intro}

In recent years, word embeddings~\cite{Mikolov2013b,Pennington2014,Bojanowski2017} have been proven to be very useful for training downstream natural language processing (NLP) tasks. Moreover, contextualized embeddings~\cite{Peters2018,devlinEtal:2019} have been shown to further improve the performance of NLP tasks such as named entity recognition, question answering, and text classification when used as word features because they are able to resolve ambiguities of word representations when they appear in different contexts. Different deep learning architectures such as multilingual BERT \cite{devlinEtal:2019}, LASER \cite{artetxeSchwenk:2019} and XLM \cite{Lample2019} have proved successful in the multilingual setting. 
All these architectures learn the semantic representations from unannotated text, making them \emph{cheap} given the availability of  texts in online multilingual resources such as Wikipedia. However, the evaluation of such resources is usually done for the high-resourced languages, where one has a smorgasbord of tasks and test sets to evaluate on. This is the best-case scenario, i.e. languages with tonnes of data for training that generate high-quality models.


For low-resourced languages, the evaluation is more difficult and therefore normally ignored simply because of the lack of resources. In these cases, training data is scarce, and the assumption that the capability of deep learning architectures to learn (multilingual) representations in the high-resourced setting holds in the low-resourced one does not need to be true. 
In this work, we focus on two African languages, \yoruba and Twi, and carry out several experiments to verify this claim. Just by a simple inspection of the  word embeddings trained on Wikipedia by fastText%
\footnote{\url{https://fasttext.cc/docs/en/pretrained-vectors.html}}, we 
see a high number of non-\yoruba or non-Twi words in the vocabularies. For Twi, the vocabulary has only 935 words, and for \yoruba we estimate that 135\,k out of the 150\,k words belong to other languages such as English, French and Arabic.

In order to improve the semantic representations for these languages, we collect online texts and study the influence of the quality and quantity of the data in the final models. We also examine the most appropriate architecture depending on the characteristics of each language.
Finally, we translate test sets and annotate corpora to evaluate the performance of both our models together with fastText and BERT pre-trained embeddings which could not be evaluated otherwise for \yoruba and Twi. The evaluation is carried out in a word similarity and relatedness task using the \emph{wordsim-353} test set, and in a named entity recognition (NER) task where embeddings play a crucial role. Of course, the evaluation of the models in only two tasks is not exhaustive but it is an indication of the quality we can obtain for these two low-resourced languages as compared to others such as English where these evaluations are already available.

The rest of the paper is organized as follows. Related works are reviewed in Section~\ref{s:sota} 
The two languages under study are described in the third section. We introduce the corpora and test sets in Section~\ref{s:data} The fifth section explores the different training architectures we consider, and the experiments that are carried out. Finally, discussion and concluding remarks are given in Section~\ref{s:conc}

\section{Related Work}
\label{s:sota}

The large amount of freely available text in the internet for multiple languages is facilitating the massive and automatic creation of multilingual resources. The resource par excellence is Wikipedia\footnote{\url{https://www.wikipedia.org}}, an online encyclopedia currently available in 307 languages\footnote{Number of languages in December 2019.}. Other initiatives such as Common Crawl\footnote{\url{https://commoncrawl.org}} or the Jehovah’s Witnesses site\footnote{\url{https://www.jw.org}} are also repositories for multilingual data, usually assumed to be noisier than Wikipedia. Word and contextual embeddings have been pre-trained on these data, so that the resources are nowadays at hand for more than 100 languages. Some examples include fastText word embeddings  \cite{Bojanowski2017,graveEtAl:2018}, MUSE embeddings~\cite{lampleEtAl:2018}, BERT multilingual embeddings \cite{devlinEtal:2019} and LASER sentence embeddings~\cite{artetxeSchwenk:2019}. In all cases, embeddings are trained either simultaneously for multiple languages, joining high- and low-resource data, or following the same methodology.

On the other hand, different approaches try to specifically design architectures to learn embeddings in a low-resourced setting. 
\newcite{ChaudharyEtAl:2018} follow a transfer learning approach that uses phonemes, lemmas and morphological tags to transfer the knowledge from related high-resource language into the low-resource one.
\newcite{jiangEtal:2018} apply Positive-Unlabeled Learning for word embedding calculations, assuming that unobserved pairs of words in a corpus also convey information, and this is specially important for small corpora.

In order to assess the quality of word embeddings, word similarity and relatedness tasks are usually used. \emph{wordsim-353}~\cite{finkelstein2001} is a collection of 353 pairs annotated with semantic similarity scores in a scale from 0 to 10. Even with the problems detected in this dataset \cite{camachoEtAt:2017}, it is widely used by the community. The test set was originally created for English, but the need for comparison with other languages has motivated several translations/adaptations. In  \newcite{hassanMihalcea:2009} the test was translated manually into Spanish, Romanian and Arabic and the scores were adapted to reflect similarities in the new language. The reported correlation between the English scores and the Spanish ones is 0.86. Later, \newcite{JoubarneInkpen:2011} show indications that the measures of similarity highly correlate across languages.
\newcite{leviantReichart:2015} translated also wordsim-353 into German, Italian and Russian and used crowdsourcing to score the pairs. Finally, \newcite{jiangEtal:2018} translated with Google Cloud the test set from English into Czech, Danish and Dutch. In our work, native speakers translate wordsim-353 into \yoruba and Twi, and similarity scores are kept unless the discrepancy with English is big (see Section~\ref{ss:eval} for details). A similar approach to our work is done for Gujarati in \newcite{JoshiEtAl:2019}.

\section{Languages under Study}
\label{s:lr}

 \paragraph{\yoruba} is a language in the West Africa with over 50 million speakers. It is spoken among other languages in Nigeria, republic of Togo, Benin Republic and Sierra Leone. It is also a language of \`{O}r{\`{i}}s{\`{a}} in Cuba, Brazil, and some Caribbean countries. It is one of the three major languages in Nigeria and it is regarded as the third most spoken native African language. There are different dialects of \yoruba in Nigeria~\cite{adegbolapatterbased16,Asahiah2014,fagbolu2015}. However, in this paper our focus is the standard \yoruba based upon a report from the 1974 Joint Consultative Committee on Education~\cite{Asahiah2017}.

Standard \yoruba has 25 letters without the Latin characters c, q, v, x and z. There are 18 consonants (b, d, f, g, gb, j[dz], k, l, m, n, p[kp], r, s, {\d s}, t, w y[j]), 7 oral vowels  (a, e, {\d e}, i, o, {\d o}, u), five nasal vowels, (an, $\underdot{e}$n, in, $\underdot{o}$n, un) and syllabic nasals 
({\`{m}}, {\'{m}}, {\`{n}}, {\'{n}}).
\yoruba is a tone language which makes heavy use of lexical tones which are indicated by the use of diacritics.  There are three tones in \yoruba namely low, mid and high which are represented as grave ({$\smallsetminus$}), macron ($-$) and acute ($/$) symbols respectively.  These tones are applied on vowels and syllabic nasals. Mid tone is usually left unmarked on vowels and every initial or first vowel in a word cannot have a high tone.  It is important to note that tone information is needed for correct pronunciation and to have the meaning of a word~\cite{adegbolaOdi12,Asahiah2014,Asahiah2017}. 
For example, \textit{ow\'{o}} (money), \textit{\d o}{\textit{w}\d {\`{o}}} (broom), \textit{\`{o}w}{\`{o}} (business), \textit{\d {\`{o}}w}{\d {\`{o}}} (honour), \textit{\d o}{w\d {\'{o}} (hand), and \textit{\d {\`{o}}w}{\d{\'{o}}}} (group) are different words with different dots and diacritic combinations. 
According to~\newcite{Asahiah2014}, Standard \yoruba uses 4 diacritics, 3 are for marking tones while the fourth which is the dot below is used to indicate the open phonetic variants of letter "e" and "o" and the long variant of "s". Also, there are 19 single diacritic letters, 3 are marked with dots below ({\d e}, {\d o}, {\d s}) while the rest are either having the grave or acute accent. The four double diacritics are divided between the grave and the acute accent as well. 

As noted in \newcite{Asahiah2014}, most of the \yoruba texts found in websites or public domain repositories \Ni either use the correct \yoruba orthography or \Nii replace diacritized characters with un-diacritized ones.
This happens as a result of many factors, but most especially to the unavailability of appropriate input devices for the accurate application of the diacritical marks \cite{adegbolapatterbased16}. This has led to research on restoration models for diacritics~\cite{orife2018attentive}, but the problem is not well solved and we find that most \yoruba text in the public domain today is not well diacritized. Wikipedia is not an exception.

\begin{table*}[ht]
\centering
\begin{tabular}{ll crr ccc}
\toprule
  Description & Source URL & \#tokens & Status & C1 & C2 & C3\\ 
\midrule
 \textit{\textbf{\yoruba}} \\
  Lagos-NWU corpus & github.com/Niger-Volta-LTI & 24,868 & clean & \cmark & \cmark &\cmark \\
  Al{\'a}k{\d {\`o}}w{\'e} & alakoweyoruba.wordpress.com & 24,092 & clean & \cmark & \cmark &\cmark\\
  {\d {\`O}}r{\d {\`o}} {\yoruba} & oroyoruba.blogspot.com & 16,232 & clean & \cmark & \cmark &\cmark\\
  {\`E}d{\`e} {\yoruba} R{\d e}w{\d {\`a}} & deskgram.cc/edeyorubarewa & 4,464 & clean & \cmark & \cmark &\cmark\\
  Doctrine \$ Covenants & github.com/Niger-Volta-LTI & 20,447 & clean & \cmark & \cmark &\cmark\\
  \yoruba Bible & www.bible.com & 819,101& clean & \cmark & \cmark &\cmark\\
  GlobalVoices & yo.globalvoices.org & 24,617& clean & \cmark & \cmark &\cmark\\
  Jehova Witness & www.jw.org/yo & 170,203& clean & \cmark & \cmark &\cmark\\
  {{\`I}r{\`i}nk{\`e}rind{\`o} n{\'i}n{\'u} igb{\'o} el{\'e}gb{\`e}je} & manual & 56,434& clean & \cmark & \cmark &\cmark\\
 Igb{\' o} Ol{\' o}d{\` u}mar{\`e} & manual & 62,125& clean & \cmark & \cmark &\cmark\\
  JW300 \yoruba corpus  & opus.nlpl.eu/JW300.php  & 10,558,055 & clean & \xmark & \xmark &\cmark\\
  \yoruba Tweets  & twitter.com/yobamoodua  & 153,716 & clean & \cmark & \cmark &\cmark\\
  BBC \yoruba & bbc.com/yoruba & 330,490& noisy & \xmark & \cmark &\cmark\\
  Voice of Nigeria \yoruba news  & von.gov.ng/yoruba & 380,252 & noisy & \xmark & \xmark &\cmark\\
  \yoruba Wikipedia & dumps.wikimedia.org/yowiki & 129,075 & noisy & \xmark & \xmark &\cmark\\
  
 \textit{\textbf{Twi}} \\
  Bible & www.bible.com & 661,229 & clean & \cmark & \cmark &\cmark\\
  Jehovah's Witness & www.jw.org/tw & 1,847,875& noisy & \xmark & \xmark &\cmark\\
  Wikipedia & dumps.wikimedia.org/twwiki & 5,820& noisy & \xmark & \cmark &\cmark\\
  JW300 Twi corpus  & opus.nlpl.eu/JW300.php  & 13,630,514  & noisy & \xmark & \xmark &\cmark \\ 
\bottomrule
\end{tabular}
\caption{Summary of the corpora used in the analysis. The last 3 columns indicate in which dataset (C1, C2 or C3) are the different sources included (see text, Section~\ref{ss:results}).}
\label{tab:corpora}
\end{table*}

 \paragraph{Twi} is an Akan language of the Central Tano Branch of the Niger Congo family of languages. It is the most widely spoken of the about 80 indigenous languages in Ghana \cite{osam}. It has about 9 million native speakers and about a total of 17--18 million Ghanaians have it as either first or second language. There are two mutually intelligible dialects, Asante and Akuapem,  and sub-dialectical variants which are mostly unknown to and unnoticed by non-native speakers. 
 It is also mutually intelligible with Fante and to a large extent Bono, another of the Akan languages. It is one of, if not the, easiest to learn to \emph{speak} of the indigenous Ghanaian languages. The same is however not true when it comes to \emph{reading} and especially \emph{writing}. This is due to a number of easily overlooked complexities in the structure of the language. 
First of all, similarly to  \yoruba, Twi is a tonal language but written without diacritics or accents. As a result, words which are pronounced differently and unambiguous in speech tend to be ambiguous in writing. Besides, most of such words fit interchangeably in the same context and some of them can have more than two meanings. A simple example is:
\begin{quote}
    Me papa aba nti na me ne wo redi no yie no. S\textipa{E} wo ara wo nim s\textipa{E} me papa ba a, me suban fofor\textipa{O} adi.
\end{quote}
This sentence could be translated as 
\begin{quote}
 \Ni   I'm only treating you nicely because I'm in a good mood. You already know I'm a completely different person when I'm in a good mood.
\end{quote}
\begin{quote}
  \Nii  I'm only treating you nicely because my dad is around. You already know I'm a completely different person when my dad comes around.
\end{quote}

Another characteristic of Twi is the fact 
that a good number of stop words have the same written form as content words. For instance, ``\emph{\textipa{E}na}'' or ``\emph{na}'' could be the words ``\emph{and, then}'', the phrase ``\emph{and then}'' or the word ``\emph{mother}''. This kind of ambiguity has consequences in several natural language applications where stop words are removed from text.

Finally, we want to point out that words can also be written with or without prefixes. An example is this same \emph{\textipa{E}na} and \emph{na} which happen to be the same word with an omissible prefix across its multiple senses.
For some words, the prefix characters are mostly used when the word begins a sentence and omitted in the middle. This however depends on the author/speaker. 
For the word embeddings calculation, this implies that one would have different embeddings for the same word found in different contexts.

\section{Data}
\label{s:data}

We collect \emph{clean} and \emph{noisy} corpora for \yoruba\ and Twi in order to quantify the effect of noise on the quality of the embeddings, where noisy has a different meaning depending on the language as it will be explained in the next subsections.

\subsection{Training Corpora}
\label{ss:corpus}

For \textbf{\yoruba}, we use several corpora collected by the Niger-Volta Language Technologies Institute%
\footnote{\url{https://github.com/Niger-Volta-LTI/yoruba-text}} with texts from different sources, including the Lagos-NWU conversational speech corpus, 
fully-diacritized \yoruba language websites and an online Bible. The largest source with clean data is the JW300 corpus.
We also created our own small-sized corpus by web-crawling three \yoruba language websites (Al{\`a}k{\d {\`o}}w{\'e}, {\d {\`O}}r{\d {\`o}} {\yoruba} and {\`E}d{\`e} {\yoruba} R{\d e}w{\d {\`a}} in Table~\ref{tab:corpora}), some Yoruba Tweets with full diacritics and also news corpora (BBC \yoruba and VON \yoruba) with poor diacritics which we use to introduce noise. By noisy corpus, we refer to texts with incorrect diacritics (e.g in BBC \yoruba), removal of tonal symbols (e.g in VON \yoruba) and removal of all diacritics/under-dots (e.g some articles in \yoruba Wikipedia). Furthermore, we got two manually typed fully-diacritized \yoruba literature ({\`I}r{\`i}nk{\`e}rind{\`o} n{\'i}n{\'u} igb{\'o} el{\'e}gb{\`e}je and Igb{\'o} Ol{\'o}d{\`u}mar{\`e}) both written by Daniel Orowole Olorunfemi Fagunwa a popular \yoruba author. 
The number of tokens available from each source, the link to the original source and the quality of the data is summarised in Table~\ref{tab:corpora}.

The gathering of clean data in \textbf{Twi} is more difficult. We use 
the Twi Bible 
as the base text as it has been shown that the Bible is the most available resource for low-resourced and endangered languages~\cite{Resnik1999TheBA}. This is the cleanest of all the text we could obtain. In addition, we use the available (and small) Wikipedia dumps which are quite noisy, 
i.e. Wikipedia contains a good number of English words, spelling errors and Twi sentences formulated in a non-natural way (formulated as L2 speakers would speak Twi as compared to native speakers). Lastly, we added text crawled from \newcite{jw} and the JW300 Twi corpus. Notice that the Bible text, is mainly written in the Asante dialect whilst the last, Jehovah's Witnesses, was written mainly in the Akuapem dialect. The Wikipedia text is a mixture of the two dialects. This introduces a lot of noise into the embeddings as the spelling of most words differs especially at the end of the words due to the mixture of dialects. The JW300 Twi corpus 
also contains mixed dialects but is mainly Akuampem. In this case, the noise comes also from spelling errors and the uncommon addition of diacritics which are not standardised on certain vowels.
Figures for Twi corpora are summarised in the bottom block of Table~\ref{tab:corpora}.

\subsection{Evaluation Test Sets}
\label{ss:eval}

\begin{table}[t]
  \centering
  \label{tab:bert_ner_result}
    \begin{tabular}{lrrrr}
    \toprule
 \textbf{Entity type} &  \multicolumn{4}{c}{\textbf{Number of tokens}} \\
 \cmidrule(l){2-5}
    & Total & Train & Val. & Test \\
 \midrule
        ORG & 289 & 214 & 40 & 35 \\
        LOC & 613 & 467 & 47 & 99 \\
        DATE & 662 & 452 & 86 & 124 \\
        PER & 688 & 469 & 109 & 110\\
        O & 23,988 & 17,819 & 2,413 & 4,867 \\
      \bottomrule
    \end{tabular}
     \caption{Number of tokens per named entity type in the Global Voices \yoruba corpus.}
     \label{tab:ner_corpus}
  \end{table}

\paragraph{\yoruba.}
One of the contribution of this work is the introduction of the wordsim-353 word pairs dataset for \yoruba. 
All the 353 word pairs were translated from English to \yoruba by 3 native speakers.
The set is composed of 446 unique English words, 348 of which can be expressed as one-word translation in \yoruba (e.g. \textit{book} translates to \textit{\`iw\'e}). In 61 cases (most countries and locations but also other content words) translations are transliterations (e.g.  \textit{Doctor} is d\'ok\'it\`a and  \textit{cucumber} is k\`uk\'umb\`a.).
98 words were translated by
short phrases instead of single words. This mostly affects  words from science and technology (e.g. \textit{keyboard} translates to p\'at\'ak\'o \`it{\d {\` e}}w\'e ---literally meaning typing board---, \textit{laboratory} translates to \`iy\`ar\'a \`i{\d s}\`ew\'ad\`i\'i ---research room---, and \textit{ecology} translates to \`im{\d {\` o}} n\'ipa \`ay\'ik\'a while \textit{psychology} translates to \`im{\d {\` o}} n\'ipa {\d {\` e}}d\'a).
Finally, 6 terms have the same form in English and \yoruba therefore they are retained like that in the dataset (e.g. Jazz, Rock and acronyms such as FBI or OPEC).

We also annotate the Global Voices \yoruba corpus to test the performance of our trained \yoruba BERT embeddings on the named entity recognition task. The corpus consists of 26\,k tokens which we annotate with four named entity types: DATE, location (LOC), organization (ORG) and personal names (PER). Any other token that does not belong to the four named entities is tagged with "O". The dataset is further split into training (70\%), development (10\%) and test (20\%) partitions. Table~\ref{tab:ner_corpus} shows the number of named entities per type and partition.

\paragraph{Twi}
Just like \yoruba, the wordsim-353 word pairs dataset was translated for Twi. Out of the 353 word pairs, 274 were used in this case. The remaining 79 pairs contain words that translate into longer phrases.

The number of words that can be translated by a single token is higher than for \yoruba. Within the 274 pairs, there are 351 unique English words which translated to 310 unique Twi words. 298 of the 310 Twi words are single word translations, 4 transliterations and 16 are used as is. 

Even if \newcite{JoubarneInkpen:2011} showed indications that semantic similarity has a high correlation across languages, different nuances between words are captured differently by languages. For instance, both \emph{money} and \emph{currency} in English translate into \emph{sika} in Twi (and other 32 English words which translate to 14 Twi words belong to this category) and \emph{drink} in English is translated as \emph{Nsa} or \emph{nom} depending on the part of speech (noun for the former, verb for the latter). 17 English words fall into this category. In translating these, we picked the translation that best suits the context (other word in the pair). In two cases, the correlation is not fulfilled at all: \emph{soap}--\emph{opera} and \emph{star}--\emph{movies} are not related in the Twi language and the score has been modified accordingly.

\begin{table*}[ht]
\centering
\footnotesize
 \begin{tabular}{p{40mm}rccrc}
    \toprule
     &  \multicolumn{2}{c}{Twi}   & & \multicolumn{2}{c}{\yoruba}  \\
      \textbf{Model} &  \textbf{Vocab Size} &  \textbf{Spearman $\rho$} && \textbf{Vocab Size} &  \textbf{Spearman $\rho$} \\
    \midrule
      \shortstack{F1: Pre-trained Model (Wiki)} & 935  & 0.143& &21,730 & 0.136 \\ \addlinespace[0.8em]
      \multirow{2}{*}[0.1em]{\shortstack{F2: Pre-trained Model~~~~~ \\ (Common Crawl \& Wiki) }} & NA & NA & &151,125 &  0.073  \\ \addlinespace[1em]
    \midrule
      \addlinespace[0.5em]
      \multirow{2}{*}[0.1em]{\shortstack{C1: Curated \textit{Small} Dataset  \\ (Clean text) }} & 9,923 & 0.354 && 12,268 &  0.322  \\ \addlinespace[1.5em]
       \multirow{2}{*}[0.1em]{\shortstack{C2: Curated \textit{Small} Dataset  \\ (Clean + some noisy text) }}& 18,494 & \textbf{0.388} & &17,492 &  0.302  \\ \addlinespace[1.5em]
      \multirow{2}{*}[0.1em]{\shortstack{C3: Curated \textit{Large} Dataset \\ (All Clean + Noisy texts) }} &  47,134 & 0.386 & & 44,560 &  \textbf{0.391} \\ \addlinespace[1.2em]
      \bottomrule
    \end{tabular}
 \caption{FastText embeddings: Spearman $\rho$ correlation between human judgements and similarity scores on the wordSim-353 for the three datasets analysed (C1, C2 and C3). The comparison with massive fastText embeddings is shown in the top rows.}
 \label{tab:fastText_result}
\end{table*}

\begin{table*}[ht]
\centering
\footnotesize
 \begin{tabular}{p{40mm}rccrc}
    \toprule
     &   \multicolumn{2}{c}{Twi}  & & \multicolumn{2}{c}{\yoruba}  \\
      \textbf{Model} &  \textbf{Vocab Size} &  \textbf{Spearman $\rho$} & &\textbf{Vocab Size} &  \textbf{Spearman $\rho$} \\
    \midrule
      \multirow{2}{*}[0.1em]{\shortstack{C1: Curated \textit{Small} Dataset  \\ (Clean text) }} & 21,819 & 0.377 & &40,162 &  0.263  \\ \addlinespace[1.5em]
       \multirow{2}{*}[0.1em]{\shortstack{C2: Curated \textit{Small} Dataset  \\ (Clean + some noisy text) }}& 22,851 & \textbf{0.437} & &56,086 &  0.345 \\ \addlinespace[1.5em]
      \multirow{2}{*}[0.1em]{\shortstack{C3: Curated \textit{Large} Dataset \\ (All Clean + Noisy texts) }} &  97,913 & 0.377 & & 133,299 &  \textbf{0.354} \\ \addlinespace[1.2em]
      \bottomrule
    \end{tabular}
\caption{CWE embeddings: Spearman $\rho$ correlation between human evaluation and embedding similarities for the three datasets analysed (C1, C2 and C3).}
\label{tab:cwe_result}
\end{table*}

\section{Semantic Representations}
\label{s:experiments}
In this section, we describe the architectures used for learning word embeddings for the Twi and \yoruba languages. Also, we discuss the quality of the embeddings as measured by the correlation with human judgements on the translated wordSim-353 test sets and by the F1 score in a NER task.

\subsection{Word Embeddings Architectures}
\label{ss:arch}
Modeling sub-word units has recently become a popular way to address out-of-vocabulary word problem in NLP especially in word representation learning~\cite{sennrich-etal-2016-neural,Bojanowski2017,devlinEtal:2019}. A sub-word unit can be a character, character $n$-grams, or heuristically learned Byte Pair Encodings (BPE) which work very well in practice especially for morphologically rich languages. Here, we consider two word embedding models that make use of character-level information together with word information: Character Word Embedding (CWE)~\cite{cwe} and fastText~\cite{Bojanowski2017}. Both of them are extensions of the Word2Vec architectures~\cite{Mikolov2013b} that model sub-word units, character embeddings in the case of CWE and character $n$-grams for fastText. 

CWE was introduced in 2015 to model the embeddings of characters jointly with words in order to address the issues of character ambiguities and non-compositional words especially in the Chinese language. A word or character embedding is learned in CWE using either CBOW or skipgram architectures, and then the final word embedding is computed by adding the character embeddings to the word itself:
\begin{equation}
    x_j = \frac{1}{2}(w_j + \frac{1}{N_j} \sum_{k=1}^{N_j} c_k)
    \label{Eqn:cwe_funtion}
\end{equation}
where $w_j$ is the word embedding of $x_j$, $N_j$ is the number of characters in $x_j$, and $c_k$ is the embedding of the $k$-th character $c_k$ in $x_j$.

Similarly, in 2017 fastText was introduced as an extension to skipgram in order to take into account morphology and improve the representation of rare words. In this case the embedding of a word also includes the embeddings of its character $n$-grams:
\begin{equation}
    x_j = \frac{1}{G_j+1} (w_j + \sum_{k=1}^{G_j} g_k)
    \label{Eqn:fasttext_funtion}
\end{equation}
where $w_j$ is the word embedding of $x_j$, $G_j$ is the number of character $n$-grams in $x_j$ and $g_k$ is the embedding of the $k$-th $n$-gram.

\newcite{cwe} also proposed three alternatives to learn multiple embeddings per character and resolve ambiguities: \Ni position-based character embeddings where each character has different embeddings depending on the position it appears in a word, i.e., beginning, middle or end \Nii cluster-based character embeddings where a character can have $K$ different cluster embeddings, and \Niii position-based cluster embeddings (CWE-LP) where for each position $K$ different embeddings are learned. We use the latter in our experiments with CWE but no positional embeddings are used with fastText.

Finally, we consider a contextualized embedding architecture, BERT~\cite{devlinEtal:2019}. BERT is a masked language model based on the highly efficient and parallelizable Transformer architecture~\cite{vaswani_transformer}  known to produce very rich contextualized representations for downstream NLP tasks. 
The architecture is trained by jointly conditioning on both left and right contexts in all the transformer layers using two unsupervised objectives: Masked LM and Next-sentence prediction. The representation of a word is therefore learned according to the context it is found in.
Training contextual embeddings needs of huge amounts of corpora which are not available for low-resourced languages such as \yoruba and Twi. However, Google provided pre-trained multilingual  embeddings for 102 languages%
\footnote{\url{https://github.com/google-research/bert/blob/master/multilingual.md}}
including \yoruba (but not Twi).

\subsection{Experiments}
\label{ss:results}

\subsubsection{FastText Training and Evaluation}

As a first experiment, we compare the quality of fastText embeddings trained on (high-quality) curated data and (low-quality) massively extracted data for Twi and \yoruba languages.

Facebook released pre-trained word embeddings using fastText for 294 languages trained on Wikipedia \cite{Bojanowski2017} (identified as F1 in Table~\ref{tab:fastText_result}) and for 157 languages trained on Wikipedia and Common Crawl \cite{graveEtAl:2018} (identified as F2 in Table~\ref{tab:fastText_result}). For \yoruba, both versions are available but only embeddings trained on Wikipedia are available for Twi. We consider these embeddings the result of training on what we call \emph{massively-extracted corpora}.
Notice that training settings for both embeddings are not exactly the same, and differences in performance might come both from corpus size/quality but also from the background model. 
The 294-languages version is trained using skipgram, in dimension 300, with character $n$-grams of length 5, a window of size 5 and 5 negatives. 
The 157-languages version is trained using CBOW with position-weights, in dimension 300, with character $n$-grams of length 5, a window of size 5 and 10 negatives. 

We want to compare the performance of these embeddings with the equivalent models that can be obtained by training on the different sources verified by native speakers of Twi and \yoruba; what we call \emph{curated corpora} and has been described in Section~\ref{s:data}
For the comparison, we define 3 datasets according to the quality and quantity of textual data used for training: \Ni \emph{Curated Small Dataset (clean), C1}, about 1.6 million tokens for \yoruba and over 735\,k tokens for Twi. The clean text for Twi is the Bible and for Yoruba all texts marked under the C1 column in Table~\ref{tab:corpora}.
\Nii In \emph{Curated Small Dataset (clean + noisy), C2}, we add noise to the clean corpus (Wikipedia articles for Twi, and BBC \yoruba news articles for \yoruba). This increases the number of training tokens for Twi to 742\,k tokens and \yoruba to about 2 million tokens. \Niii \emph{Curated Large Dataset, C3} consists of all available texts we are able to crawl and source out for, either clean or noisy. The addition of JW300~\cite{agic-vulic-2019-jw300} texts increases the vocabulary to more than 10\,k tokens in both languages.

We train our fastText systems using
a skipgram model with an embedding size of 300 dimensions, context window size of 5, 10 negatives and $n$-grams ranging from 3 to 6 characters similarly to the pre-trained models for both languages. Best results are obtained with  minimum word count of 3.
Table~\ref{tab:fastText_result} shows the Spearman correlation between human judgements and cosine similarity scores on the wordSim-353 test set. Notice that pre-trained embeddings on Wikipedia show a very low correlation with humans on the similarity task for both languages ($\rho$=$0.14$) and their performance is even lower when Common Crawl is also considered ($\rho$=$0.07$ for \yoruba).
An important reason for the low performance is the limited vocabulary. The pre-trained Twi model has only 935 tokens. For \yoruba, things are apparently better with more than 150\,k tokens when both Wikipedia and Common Crawl are used but correlation is even lower. An inspection%
\footnote{We used \textit{langdetect} to have a rough estimation of the language of each word, assuming that words that are not detected are \yoruba because the language is not supported by the tool.}  of the pre-trained embeddings indicates that over 135\,k words belong to other languages mostly English, French and Arabic. 
If we focus only on Wikipedia, we see that many texts are without diacritics in \yoruba and often make use of mixed dialects and English sentences in Twi.

The Spearman $\rho$ correlation for fastText models on the curated small dataset (clean), C1, improves the baselines by a large margin ($\rho=0.354$ for Twi and 0.322 for \yoruba) even with a small dataset. The improvement could be justified just by the larger vocabulary in Twi, but in the case of \yoruba the enhancement is there with almost half of the vocabulary size.
We found out that adding some noisy texts (C2 dataset) slightly improves the correlation for Twi language but not for the \yoruba language. The Twi language benefits from Wikipedia articles because its inclusion doubles the vocabulary and reduces the bias of the model towards religious texts. However, for \yoruba, noisy texts often ignore diacritics or tonal marks which increases the vocabulary size at the cost of an increment in the ambiguity too. As a result, the correlation is slightly hurt. One would expect that training with more data would improve the quality of the embeddings, but we found out with the results obtained with the C3 dataset, that only high-quality data helps. The addition of JW300 boosts the vocabulary in both cases, but whereas for Twi the corpus mixes dialects and is noisy, for \yoruba it is very clean and with full diacritics. Consequently, the best embeddings for \yoruba are obtained when training with the C3 dataset, whereas for Twi, C2 is the best option. In both cases, the curated embeddings improve the correlation with human judgements on the similarity task a $\Delta\rho=+0.25$ or, equivalently, by an increment on $\rho$ of 170\% (Twi) and 180\% (\yoruba).

\subsubsection{CWE Training and Evaluation}
\label{ss:cwe}

The huge ambiguity in the written Twi language motivates the exploration of different approaches to word embedding estimations. In this work, we compare the standard fastText methodology to include sub-word information with the character-enhanced approach with position-based clustered embeddings (CWE-LP as introduced in Section~\ref{ss:arch}). With the latter, we expect to specifically address the ambiguity present in a language that does not translate the different oral tones on vowels into the written language.

The character-enhanced word embeddings are trained using a skipgram architecture with cluster-based embeddings and an embedding size of 300 dimensions, context window-size of 5, and 5 negative samples. In this case, the best performance is obtained with a minimum word count of 1, and that increases the effective vocabulary that is used for training the embeddings with respect to the fastText experiments reported in Table~\ref{tab:fastText_result}.

We repeat the same experiments as with fastText and summarise them in Table~\ref{tab:cwe_result}.
If we compare the relative numbers for the three datasets (C1, C2 and C3) we observe the same trends as before: the performance of the embeddings in the similarity task improves with the vocabulary size when the training data can be considered clean, but the performance diminishes when the data is noisy.

According to the results, CWE is specially beneficial for Twi but not always for \yoruba. Clean \yoruba text, does not have the ambiguity issues at character-level, therefore the $n$-gram approximation works better when enough clean data is used ($\rho^{C3}_{CWE}=0.354$ vs. $\rho^{C3}_{fastText}=0.391$) but it does not when too much noisy data (no diacritics, therefore character-level information would be needed) is used ($\rho^{C2}_{CWE}=0.345$ vs. $\rho^{C2}_{fastText}=0.302$). For Twi, the character-level information reinforces the benefits of clean data and the best correlation with human judgements is reached with CWE embeddings ($\rho^{C2}_{CWE}=0.437$ vs. $\rho^{C2}_{fastText}=0.388$).

\subsubsection{BERT Evaluation on NER Task}
\label{ss:bert}

\begin{table}[t]
  \centering
  \footnotesize
  \label{tab:bert_ner_result}
  \scalebox{0.85}{
    \begin{tabular}{p{30mm}p{7mm}p{7mm}p{7mm}p{7mm}r}
    \toprule
      \textbf{Embedding Type} &  \textbf{DATE} &   \textbf{LOC} &  \textbf{ORG} &  \textbf{PER}  & \textbf{F1-score} \\
    \midrule
      \multirow{2}{*}[0.2em]{\shortstack{Pre-trained \textit{uncased} \\ Multilingual-bert \\ (Multilingual vocab)}} & 44.6 & 33.9 & 12.1 & 5.7 &  $27.1 \pm 0.7$ \\ \addlinespace[3.0em]
       \multirow{2}{*}[0.2em]{\shortstack{Fine-tuned \textit{uncased} \\ Multilingual-bert \\(Multilingual vocab) }} & 64.0 & 65.3 & 38.8 & 47.4 & $56.4 \pm 2.4$ \\ \addlinespace[3.0em]
      \multirow{2}{*}[0.2em]{\shortstack{Fine-tuned \textit{uncased} \\ Multilingual-bert \\ (\yoruba vocab)}} & 67.0 & 71.5 & 40.4 & 49.4 &  $60.1 \pm 0.8$ \\ \addlinespace[2.0em]
      \bottomrule
    \end{tabular}
    }
     \caption{NER F1 score on Global Voices \yoruba corpus after fine-tuning BERT for 10 epochs. Mean F1-score computed after 5 runs}
     \label{tab:ner}
  \end{table}

In order to go beyond the similarity task using static word vectors, we also investigate the quality of the multilingual BERT embeddings by fine-tuning a named entity recognition task on the \yoruba Global Voices corpus. 

One of the major advantages of pre-trained BERT embeddings is that fine-tuning of the model on downstream NLP tasks is typically computationally inexpensive, often with few number of epochs. However, the data the embeddings are trained on has the same limitations 
as that used in massive word embeddings.
Fine-tuning involves replacing the last layer of BERT used optimizing the masked LM with a task-dependent linear classifier or any other deep learning architecture, and training all the model parameters end-to-end. For the NER task, we obtain the token-level representation from BERT and train a 
conditional random field classifier for sequence tagging.

Similar to our observations with non-contextualized embeddings, we find out that fine-tuning the pre-trained multilingual-uncased BERT for 10 epochs on the NER task gives an F1 score of 27. If we do the same experiment in English, F1 is 66.2 after 10 epochs.
That shows how pre-trained embeddings by themselves do not perform well in downstream tasks on low-resource languages. To address this problem for \yoruba, we fine-tune BERT 
masked language model on the \yoruba corpus in two ways: \Ni using the multilingual vocabulary, and \Nii using only \yoruba vocabulary. In both cases diacritics are ignored to be consistent with the base model training.

As expected, the fine-tuning of the pre-trained BERT on the \yoruba corpus in the two configurations generates better representations than the base model. These models are able to achieve a better performance on the NER task with an average F1 score of over 56\% (see Table~\ref{tab:ner} for the comparative). The fine-tuned BERT model with only \yoruba vocabulary further increases by 
4\% in F1 score than the BERT model that uses the multilingual vocabulary. 
Although we do not have enough data to train BERT from scratch, we observe that fine-tuning BERT on a limited amount of monolingual data of a low-resource language helps to improve the quality of the embeddings. The same observation holds true for high-resource languages like German\footnote{\url{https://deepset.ai/german-bert}} and French~\cite{martin2019camembert}.

\section{Summary and Discussion}
\label{s:conc}

In this paper, we present curated word and contextual embeddings for \yoruba and Twi. For this purpose, we gather and select corpora and study the most appropriate techniques for the languages. We also create test sets for the evaluation of the word embeddings within a word similarity task (wordsim353) and the contextual embeddings within a NER task. Corpora, embeddings and test sets are available in github%
\footnote{\url{https://github.com/ajesujoba/YorubaTwi-Embedding}}.

In our analysis, we show how massively generated embeddings perform poorly for low-resourced languages as compared to the performance for high-resourced ones.
This is due both to the quantity but also the quality of the data used. While the Pearson $\rho$ correlation for English obtained with fastText embeddings trained on Wikipedia (WP) and Common Crawl (CC) are $\rho_{WP}$=$0.67$ and $\rho_{WP+CC}$=$0.78$, the equivalent ones for \yoruba are $\rho_{WP}$=$0.14$ and $\rho_{WP+CC}$=$0.07$. For Twi, only embeddings with Wikipedia are available ($\rho_{WP}$=$0.14$). By carefully gathering high-quality data and optimising the models to the characteristics of each language, we deliver embeddings with correlations of $\rho$=$0.39$ (\yoruba) and $\rho$=$0.44$ (Twi) on the same test set, still far from the high-resourced models, but representing an improvement over $170\%$ on the task.

In a low-resourced setting, the data quality, processing and model selection is more critical than in a high-resourced scenario. We show how the characteristics of a language (such as diacritization in our case) should be taken into account in order to choose the relevant data and model to use. As an example, Twi word embeddings are significantly better when training on 742\,k selected tokens than on 16 million noisy tokens, and when using a model that takes into account single character information (CWE-LP) instead of $n$-gram information (fastText).

Finally, we want to note that, even within a corpus, the quality of the data might depend on the language. Wikipedia is usually used as a high-quality freely available multilingual corpus as compared to noisier data such as Common Crawl. However, for the two languages under study, Wikipedia resulted to have too much noise: interference from other languages, text clearly written by non-native speakers, lack of diacritics and mixture of dialects. The JW300 corpus on the other hand, has been rated as high-quality by our native \yoruba speakers, but as noisy by our native Twi speakers. In both cases, experiments confirm the conclusions.

\section{Acknowledgements}
The authors thank Dr. Clement Odoje of the Department of Linguistics and African Languages, University of Ibadan, Nigeria and Ol{\'o}y{\`e} Gb{\'e}mis{\'o}y{\`e} {\`A}r{\d{\`{e}}}{\'o} for helping us with the \yoruba translation of the WordSim-353 word pairs and Dr. Felix Y. Adu-Gyamfi and Ps. Isaac Sarfo for helping with the Twi translation.  We also thank the members of the Niger-Volta Language Technologies Institute for providing us with clean \yoruba corpus

The project on which this paper is based was partially funded by the German Federal Ministry of Education and Research under the funding code 01IW17001 (Deeplee). Responsibility for the content of this publication is with the authors.

\section{Bibliographical References}
\label{main:ref}

\bibliographystyle{lrec}
\bibliography{embeddingsAfrican}

\end{document}